# ANUSAARAKA: MACHINE TRANSLATION IN STAGES

*Akshar Bharati*
*Vineet Chaitanya*
*Amba P. Kulkarni*
*Rajeev Sangal*

**IIT Kanpur Centre for NLP at Hyderabad**
**Central University Campus**
**Hyderabad**
*{vineet,amba,sangal}@uohyd.ernet.in*

**1.   INTRODUCTION:**

Fully-automatic general-purpose high-quality machine translation systems (FGH-MT) are extremely difficult to build.  In fact, there is no system in the world for any pair of languages which qualifies to be called FGH-MT.  The reasons are not far to seek.  Translation is a creative process which involves interpretation of the given text by the translator.  Translation would also vary depending on the audience and the purpose for which it is meant.  This would explain the difficulty of building a machine translation system.  Since, the machine is not capable of interpreting a general text with sufficient accuracy automatically at present - let alone re-expressing it for a given audience, it fails to perform as FGH-MT.  FOOTNOTE{The major difficulty that the machine facesin interpreting a given text is the lack of general world knowledge or common sense knowledge.}

To understand the nature of the difficulty, let us consider the following sentence in Hindi:

```
 chAvala  rAma     khAtA  hE
 rice(m.) Ram(m.)  eats(m.)
 Ram eats rice.
```

in which the karta (or the eater) is not marked.  Translation to English, say, would be quite different depending on which one is the karta. (The problem is that there are no vibhakits (post-positions or case endings) to indicate which one is the karta or 'eater'. Noun-verb agreement normally helps in identifying the karta because the gender-number-person (gnp) of the verb agrees with the karta.  Here, it does not help us in identifying the karta unambiguously because the masculine (m.) ending of the verb tells us that the karta is masculine, and there are two masculine nouns (Ram or rice).)

It seems trivial for us to assume that a person (Ram, in this sentence) would be the agent of eating, and rice, the thing which is eaten.  But the machine does not "know" that.  This knowledge is said to be world knowledge, as it pertains to the world as it exists.  It turns out, that if we try to put this kind of information in the mchine, we find that there are a very large number of such facts.  For all the nouns, we will have to say who can eat whom.  How should such facts be organized, is the first problem?  But there is a still harder problem that turns up.  Such knowledge is quite easily over-ridden in language to convey a metaphorical sense, irony, etc. Consider the following sentence:

    Apa SarAba  nahiM pIte,  SarAba  Apako          pItI hE  
    You alcohol not   drink, alcohol you (accus.) drinks.  
    You do not drink alcohol, alcohol drinks you.

Here, the sentence says that the alcohol drinks a person, and yet it is a perfectly good sentence!

Thus, it is not enough to put such a large number of facts about the world in the machine, we must also put conditions regarding when they can be overridden while processing text.  This turns out to be an incredibily hard task.

    The examples we have considered, are rather easy because the world knowledge that needs to be referred to is fairly well pin-pointed.  Quite frequently, there is ellipses in sentences (i.e. parts of sentences are dropped).  The missing part may have to be inferred (possibly using world knowledge) before the processing can be done further.  Similarly many times the chain of reasoning is very long and might require a deep understanding of the preceding text.

    Does it mean that since the machine cannot interpret text with a fair degree of accuracy, machine translation must be abandoned as a distant dream?  The answer lies in sharing the load between the reader and the machine so that the tasks which are hard for the human being are done by the machine and vice versa, as explained below.  This is discussed in the next two sections.

## 2. Solution

Different researchers have tried to give different answers to the problem.  The most common approach has been to delimit the subject domain so that machine works in a narrow subject area, such as, weather reports aircraft maintenance manuals, computer manuals, etc.  It has been hoped that by delimiting MT in a narrow area, one stands  a better chance of using context, domain knowledge, etc.  The system would perform badly when give a text outside the domain  but that is a limitation one would have to live with.  The real difficulty is in identifying  a domain that is narrow enough that the system works well, and wide enough that enough real texts qualify to be in it, so that it is practically useful.

The anusaaraka answer is different from the above.  Here the task of building an MT System is subdivided into two parts;

(1) The first module (called core anusaaraka) does language-based analysis: It takes all the information in the source text and presents it in its output, in an intermediate language which is quite close to the target language.

(2) The second module may do  domain specific knowelge based processing, statistical processing, etc. in which it may utilize world knowledge, frequency information, concordances, etc. to produce output in the target language.

There are many advantages of the above approach:

(1)     Early availability:  The first module is much easier to build than the second, requiring an order of magnitude less effort.  Therefore, the system becomes available for use at an early date. However, it must be pointed out that a certain amount of training would be needed for the user, to read the output in the intermediate language.

(2) Early Feedback: Since the system can be used at an early date, not only does it serve a useful purpose, it also becomes easier to build the second module. The early feedback guides the refinement and building of the system .

(3) Robustness: The second module by its very nature will be fragile. The first module will be much more rebust. Therefore, the system provides a robust layer which can be used even if the second module fails in any specific text.

(4) Clear task demarcation: By separating the two modules based on clear cut criteria (of language knowledge and background knowledge), it becomes a lot easier to coordinate the activities of system building. The builders of each module understand the limits within which they have to operate, therefore, it becomes possible to engage in a large coordinated team effort.

(5) Evolution of interlingua: The output language would have some constructions of the source language, those which cannot be transferred to any existing construction in the target language (and hence requires training as mentioned earlier). By using this approach with many source and target languages, an interlingua would evolve naturally. Thus, the intermediate language no longer needs to be designed beforehand, rather it evolves naturally.

(6) The second module becomes easier to build once the first module is ready because a person working on the second module does not need to know two languages (source and target). He needs to know only one, namely, the intermediate language, which is close to the target language.

(7) The anusaaraka approach can also be used in contexts where translation is inappropriate. For example, if a law book is being studied for examining interpretations of the law, clearly translation is of no use. The original needs to be read faithfully and examined carefully. Here, the anusaaraka (namely, only the first module) is what is needed. Similar would be the case when a scholar is studying a classic trying to see what the author has actually written. Thus, anusaaraka is an independent concept distinct from translation which can have its own independent use.

**3. Anusaaraka Approach**

Anusaaraka maps constructions in the source language to the corresponding constructions in the target language wherever possible. For example, a noun or pronoun in the source language is mapped to an approriate noun or pronoun, respectively, in the target language below:

```
 T: mIru  pustakaM  caduvutunnArA?
@H: Apa   pustaka   paDha_raHA_[HE|thA]_kyA{23_ba.}?
!E: You   book      read_ing_[is|was] Q.?
 E: Are/were you reading a book?
```

(Where the prefixes mean the following:
T=Telugu, @H=anusaaraka Hindi, !E=English gloss, E=English.)

In the example above, the last word in the sentence is a verb and illustrates the mapping from Telugu to Hindi morpheme by morpheme: the root is mapped to 'paDha' (read), and similarly the tense-aspect-modality (TAM) lable is mapped to 'raHA_[HE|thA]' (is_*ing or was_*ing), which is followed by 'A' sufix which gets mapped to 'kyA' (what) as a question marker in Hindi. Telugu leaves the tense open as present or past, which is reflected in the output. Gender, number, and person (GNP) information is is also shown separately in curly brackets ('{23_ba.}' for second or third person and bahu-vachana (plural)).

Sometimes, for a construction in the source language, the same construction is not available in the target language. In such a case, the system choses another construction in the target language in which the same information can be expressed. In the example below, the system choses

the complementizer construction in Hindi (EsA) to express the same sense:

```
 T: mA      ammAyiki    uxyogaM ceVyyAlani    lexu.
@H: hamArA_ ladakI_ko`  nOkarI  karanA_EsA    nahIM_[hE|WA].
!E: Our     daughter(dat.) job  do_should_that not(fem.)
 E: It is not the case that our daughter should get a job.
```

However, the anusaaraka shows the image and therefore, it uses the complementizer (EsA).

Sometimes there are slight differences between a construction in the source language to a similar construction in the target language because of which information might not be preserved. In such a situation additional notation is introduced to express the information which would otherwise get lost. A simple example of this is the lack of distinction between personal pronoun and pronominal adjective in Hindi: vaha. Telugu on the other hand has two different words for these: vADu and A.

```
 T: vADu   shkUluku    vellADu.
@H: vaha`  pAThshAlA_ko` gayA.
!E: he     school(dat.) went.
 E: He went to school.

 T: A      pAThaSAlaku  TrophI vacciMdi.
@H: vaha-  pAThshAlA_ko` TrophI AyI.
!E: that   school(dat.) trophy came
 E: That school received the trophy.
```

When transferring from Telugu to Hindi, this distinction would have disappeared, if care was not taken. In anusaaraka, the two forms are made different by introducing additional notation:

```
 vaha`   (he)
 vaha-   (that)
```

A more complex example involving both the use of additional notation and mapping to a different construction is the participial construction in South Indian languages. This is an example of building "language-bridges". In South Indian languages, there are a rich set of participial verbs. Hindi has basically only two. Therefore, to map all the participial verbs from these languages, the relative clause construction in Hindi is used.

```
 T: pani cesina      rAmmUrti    maMcivADu.
@H: kAma kiyA_HE_jo*_vaHa_ rAmmUrti_[-] bhalA_AdamI[acchA_{pu.}].
!E: work *who_has_done Ramamurti    good_man
 E: Ramamurti who has done the word is a good man.
```

It should be noted above that the participial construction in Telugu ('cesina' as a modifier of the following noun 'rAmmUrti') is mapped to a relative clause construction in Hindi. However, there is some information which is not present in the original Telugu construction, but is required in the relative clause construction in Hindi (marked by '*' above). This can be supplied either by the user or by the second module based on a suitable knowledge base.  In this case, 'ne' should be put in Hindi for the doer of the work; in general, it could be 'ko' for the karma, 'se' for the instrument etc.

However, it should be rememberd that the second module would be fragile and will fail at times. In those situations, the anusaaraka output can be read by a reader to get the correct meaning.

There can be other failures from unexpected quarters. Here is an example:

```
 T: UriniMci   oVccina        nalini      mA     ceVlleVlu.
 @H: gAzva_se`  AyA_hE_jo*_vaha_  bImArI_ko[hUz] hamArA_ CotI_bahana.
              ^0
!E: from_village *who_has_come   illness_acc.  my     younger_sister.
         ^one_
 E: Nalini, who has come from the village is my younger sister.
```

Here, 'nalini' a proper noun got translated to 'illness'. This failure can cause difficulty for the second module if it tries to arrive at a suitable replacement for '*', and the reader can fall back on the core anusaaraka to infer the meaning.

### 4. Conclusion

We have described the anusaaraka approach  in which the task of building an MT system is cleanly broken into two modules:
(1) the core anusaaraka systsem which is based on language knowledge, and
(2) the domain specific module based on world knowledge, statistical knowledge, etc.

The core anusaaraka output is in a language close to the target language, and can be understood by the human reader after some training.

The question is how much training is necessary to get a very high degree of comprehension. Our experience of working among Indian languages shows that this training is likely to be small. Reason for this is that India forms a linguistic area: Indian languages share vocablulary and grammatical constructions. There are also shared pragmatics and culture.  Similar approach can be applied to build an English to Hindi anusaaraka. A study can be conducted related to training required to read such an output. The expectation is that an English to Hindi usable system can be built except that it will require longer training.